\documentclass[journal]{IEEEtran}
\ifCLASSINFOpdf
\else
\fi
\usepackage[noadjust]{cite}
\usepackage{amsmath,amssymb,amsfonts}
\usepackage{algorithmic}
\usepackage{graphicx}
\usepackage{textcomp}
\usepackage{xcolor}
\usepackage{xr} % able to refer to labels from other documents
\usepackage{booktabs}
\usepackage[colorlinks=true,linkcolor=blue, citecolor=blue]{hyperref}

\newcommand{\ra}[1]{\renewcommand{\arraystretch}{#1}}
\usepackage{mathtools}
\DeclarePairedDelimiter{\ceil}{\lceil}{\rceil}
\def\BibTeX{{\rm B\kern-.05em{\sc i\kern-.025em b}\kern-.08em
    T\kern-.1667em\lower.7ex\hbox{E}\kern-.125emX}}

\begin{document}
\title{Neural Networks as Geometric Chaotic Maps}


\author{Ziwei~Li,~
        and~Sai~Ravela%
\thanks{Ziwei Li and Sai Ravela are with the Department
of Earth, Atmospheric, and Planetary Sciences, Massachusetts Institute of Technology, Cambridge,
MA, 02139 (e-mail: ziweili@mit.edu, ravela@mit.edu).}}


\markboth{\MakeLowercase{submitted to} IEEE Transactions on Neural Networks and Learning Systems}%
{Shell \MakeLowercase{\textit{et al.}}: Bare Demo of IEEEtran.cls for IEEE Journals}



\maketitle

\begin{abstract}
The use of artificial neural networks as models of chaotic dynamics has been rapidly expanding. Still, a theoretical understanding of how neural networks learn chaos is lacking. Here, we employ a geometric perspective to show that neural networks can efficiently model chaotic dynamics by becoming structurally chaotic themselves. We first confirm neural network's efficiency in emulating chaos by showing that a parsimonious neural network trained only on few data points can reconstruct strange attractors, extrapolate outside training data boundaries, and accurately predict local divergence rates. We then posit that the trained network's map comprises sequential geometric stretching, rotation, and compression operations. These geometric operations indicate topological mixing and chaos, explaining why neural networks are naturally suitable to emulate chaotic dynamics.
\end{abstract}


% \fi
\IEEEpeerreviewmaketitle

\section{Introduction}
\IEEEPARstart{C}{haotic} dynamics are ubiquitous in observed and simulated trajectories of physical systems\;\cite{Strogatz2015}. 
Finding the exact solutions to such dynamics is often impossible due to the nonlinearities in the system equations and the characteristic exponential divergence from two initially close-by  trajectories. 
In the absence of first-principle theories, data-driven low-dimensional models are often implemented to capture the observed behavior, from which theoretical understanding may emerge. 
When theories are explicitly available, modelers may numerically simulate the governing equations. Using an ensemble simulation with randomly perturbed initial conditions, they quantify errors, uncertainty, and predictability in the solutions. 
However, doing so is often challenging because the simulations are high-dimensional (thus expensive), the equations are nonlinear, and the uncertainties are non-Gaussian\;{\cite{Ravela2018}}. 
In both the data-driven and numerical modeling frameworks, approximate low-complexity schemes are sought to predict and understand chaotic systems. 
Devising efficient models of chaotic systems is thus an important subject in engineering and physical sciences.

The emergence of artificial neural networks and the associated universal approximation theorem (UAP)\;{\cite{Hornik1989, Hornik1991, Seidl1991, Funahashi1993}} suggest that the neural network is a generalized modeling tool. It would seem that, provided with enough neurons, they can emulate any dynamical system of any complexity. 
Indeed, a surge of interest in using neural networks to emulate chaotic systems has emerged\;{\cite{Garliauskas1998, Bakker2000, Das2000, Das2002, Dudul2005, Yang2006, Bahi2012, Song2012, Zerroug2013, Prezioso2015, Zhang2017, Zhang2017b, Yu2017, Pathak2017, Madondo2018, Chen2018, Pathak2018, Bao2020}}, with additional theoretical development using them  to control nonlinear dynamical systems \cite{Tan2020, Liu2020, Liu2020a, Tan2021}. 

Prior work using neural networks to model chaotic dynamics can be divided into two categories, 
a) designing simple networks that achieve chaotic behavior (top-down) and b) learning from data generated by detailed simulation or observation (bottom-up). 

Numerous top-down approaches through numerical simulations and hardware-programming already show that simple neural networks become chaotic in specific parameter regimes. 
Models of 3 or 5 neurons with a truncated polynomial activation function, for example, show chaotic behaviors\;\cite{Garliauskas1998}. Likewise, 
3D \;\cite{Das2000, Das2002} and 2D\;\cite{Zerroug2013} neural maps show period-doubling bifurcations to give onset of chaos.
A 3D cellular network is shown to obtain a horseshoe map for some weight matrices\;\cite{Yang2006}. 
A delayed 2D network depicts chaotic dynamics\;\cite{Song2012}, and a 3D variant of memristor circuits inspired by biological neuron firings\;\cite{Prezioso2015} is  found to be chaotic through the fold and Hopf bifurcations\;\cite{Bao2020}. 

The bottom-up approach addresses using artificial neural networks to learn from data to reproduce chaotic dynamics. 
Bakker et al. (2000) \cite{Bakker2000} use multilayer perceptron models in addition to linear models to propagate embedded data from a chaotic pendulum. Bahi et al. (2012) \cite{Bahi2012} establish the equivalence between a class of neural networks and Devaney's definition of chaos. The dynamics of the classic Lorenz-63 system \cite{Lorenz1963} can be emulated with simple feedforward networks\;\cite{Zhang2017, Zhang2017b}, with recurrent networks\;\cite{Yu2017}, and with LSTM~\cite{Madondo2018}. Neural networks are also thought of as nonlinear ODE propagators~\cite{Chen2018, Trautner2020} and find  use in quantifying Lyaponuv exponents in higher dimensional chaotic systems \cite{Pathak2017, Pathak2018}.

In contrast to previous research, this paper explains how trained neural networks become chaotic. We refer to ``neural networks" as the traditional static feedforward neural network (NN). The novelty of our work is that:
\begin{itemize}
\item We use the finite-time Lyapunov exponent as a predictability measure. It reveals that the dynamics of NN is ``as chaotic as" the true system at both short and long timescales. 
\item We propose a novel geometric perspective, which explains how trained NNs are able to efficaciously model chaos
 at very low complexity (number of neurons). 
\end{itemize}
We illustrate NN's efficacy in emulating chaos by training a parsimonious single-hidden-layer NN on the Lorenz-63 system (L63). This NN reconstructs the L63 attractor structure with well-matched predictability and very high fidelity. Training with a one-sided segment of the attractor also delivers well-behaved trajectories on the other side, indicating that the trained NN extrapolates far beyond the training set. 

To the best of our knowledge, no existing theory seems to explain why chaos appears {\it learnable}{\footnote{``Learnable"  is used in the sense of a matched predictability between the true dynamical system and NN, quantified by finite-time Lyapunov exponents. It is different from, e.g., Valiant's definition\;{\cite{Valiant1984}}.}} to such simple neural networks. Unfortunately, the universal approximation theorem is of little help because it neither explains the emergence of chaos in NN nor the efficacy with which it emerges. 

Instead, we posit through a new geometric view that the trained network's flow induces topological mixing, 
which explains how chaos develops in NN. 
The NN flow alternately rotates, stretches, and compresses in phase space, 
which are the defining characteristics of chaotic dynamics\;{\cite{Berge1987}}. 
Tucker (2002) {\cite{Tucker2002}} also uses a similar view to devise a geometric Lorenz map to explain the topological properties of L63. 
These geometric transformations required by chaos theory enable the effective reconstruction of strange attractors by NN while matching the true system's predictability. 
Simplicity in NN's structure is central in the effectiveness argument. Using a new, tighter bound on NN's complexity, we show that simple NN in fact learns from chaotic systems with great efficacy.  

The remainder of this paper is organized as follows. Section \ref{sec:neu} reports experimental results about the learnability of NN on chaotic dynamics using L63. Section \ref{sec:geo} explains the efficacy of NN from a geometrical perspective. We then provide bounds on the complexity of neural networks in section \ref{sec:bounds} and conclude in \ref{sec:concl}.

\section{Neural Lorenz-63 emulation}
\label{sec:neu}

The L63 model originally describes the 2D Rayleigh-B\'enard convection. Truncating the spectral components of the dynamical fields yields a set of ordinary differential equations \cite{Lorenz1963}: 
\begin{equation}
\begin{aligned}
    \dot{X} &= \sigma(Y - X),\\
    \dot{Y} &= \rho X - Y - XZ,\\
    \dot{Z} &= -\beta Z + XY, 
\end{aligned}
\label{eq:L63}
\end{equation}
where $X$ and $Y$ are the magnitudes of the stream function and temperature modes, and $Z$ is the deviation of the vertical temperature profile from linearity. Consistent with typical applications {\cite{Lorenz1963}}, we set $\sigma = 10$, $\beta = 8/3$, and $\rho = 28$. The solutions of L63 are known to be dissipative (volume in phase space contracts rapidly) and chaotic (sensitive to initial perturbations). The discrete L63 map describes a map from the current state of the system $\mathbf{x}_n = (X, Y, Z)^{\mathrm{T}}$ to the state at the next timestep $\mathbf{x}_{n + 1}$
\begin{equation}
\begin{aligned}
\mathbf{\Phi}_{\text{L63}}(\mathbf{x}_n) \mapsto \mathbf{x}_{n + 1}. 
\label{eq:mapping_discrete}
\end{aligned}
\end{equation}
We analyze the discrete maps of L63 and NN because they provide a direct geometric connection between the dynamics of L63 and NN, as we shall discuss in section \ref{sec:geo}. Since the analytical form of {\eqref{eq:mapping_discrete}} is unknown, the discrete map of L63 is obtained by numerically integrating {\eqref{eq:L63}} with a uniform time step $\mathrm{d}t = 0.01$. 

\subsection{Compact neural model}
\label{sec:compactNN}

We use single-hidden-layer feedforward neural networks to learn the dynamics of L63. 
Networks with more hidden layers are no doubt feasible, but we seek the simplest network which is easy to interpret geometrically and also agrees well with the numerical solution. 
The functional form of an $L$-neuron NN map is: 
\begin{equation}
\mathbf{\Phi}_{\text{NN}}(\mathbf{x}_n) = \mathbf{W}_2g(\mathbf{W}_1\mathbf{x}_n + \mathbf{b}_1) + \mathbf{b}_2, 
\label{eq:NN}
\end{equation}
where $\mathbf{x}_{n}$ is the $3\times1$ input vector, $\mathbf{W}_1$ is an $L\times3$ weight matrix, $\mathbf{b}_1$ is an $L\times1$ bias term, and $g(\cdot)$ is the activation function. An $L\times 3$ weight matrix $\mathbf{W}_2$ and an $L\times 1$ bias $\mathbf{b}_2$ connect the hidden layer to the output. 

The discrete map of L63 is solved with Matlab function \textit{ode45} to generate training data. To obtain data on the attractor, we randomly initialize 1000 trajectories from region $[-20, 20]\times[-20, 20]\times[0, 50]$ with uniform distribution. Each trajectory is integrated for 2500 timesteps. We abandon the first 2000 timesteps to remove the transient parts (typically much shorter than 2000 steps). The remaining 500 timesteps of the 1000 trajectories are aggregated as pairs $(\mathbf{x}, \mathbf{x'})$ that satisfy $\mathbf{x'} = \mathbf{\Phi}_{L63}(\mathbf{x})$ to form the training data pool. The ensemble of $\mathbf{x}$ represents the L63 attractor ($\mathcal{A}_{\text{L63}}$), and the ensemble of location pairs provides information about the L63 flow. We randomly sample 20, 30, 40, 60, 100, 150 location pairs from the training data pool and train single-hidden-layer NNs with 3-8 neurons. Each NN is trained for $10^3$ epochs with Bayesian regularization\;{\cite{DanForesee1997}}. 

\begin{figure}[ht]
    \centering
    \includegraphics[width=0.8\linewidth]{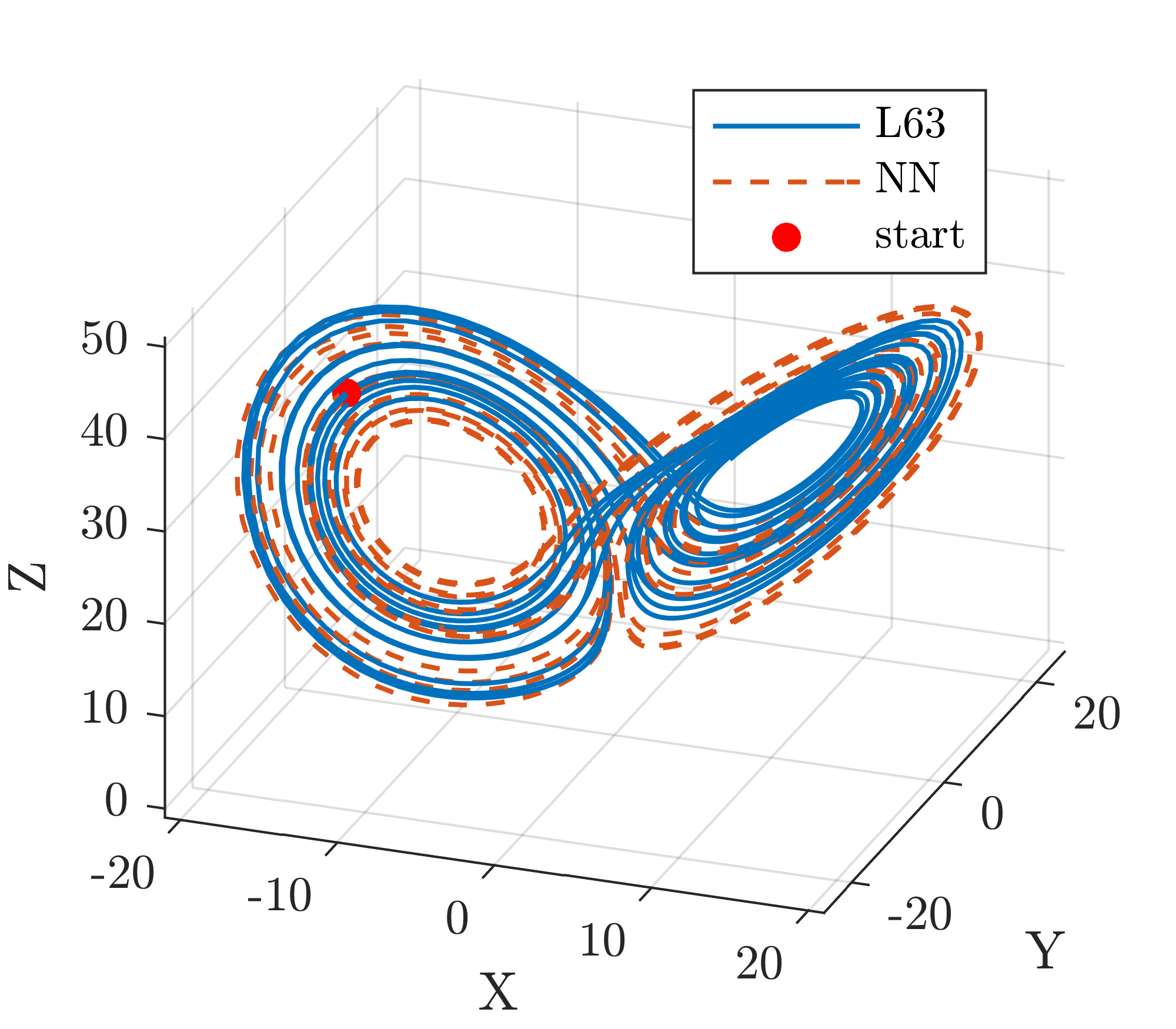}
\caption{Two trajectories produced by L63 (blue) and the 4-neuron NN trained on 40 data points sampled from the whole attractor (red-dashed). They start from the same location on the L63 attractor (red dot), and are both 2000 timesteps long.}
\label{fig:NN_predictions_small_num}
\end{figure}

Here, $\tanh$ is chosen as the activation function, $g(\cdot)$, for the rest of the paper, which is representative for the necessary nonlinearity in the neural networks. 
For completeness, we find that the choice of $g(\cdot)$ matters for the performance of the NN 
%(Fig.\;\ref{fig:errors}). 
(Fig.\;S2). 
Nonlinear sigmoidal functions such as Elliot sigmoid and log sigmoid have a similar performance to $\tanh$, most likely due to their similar functional forms. 
Nonlinear radial basis function ($e^{-|x|^2}$) and softplus [$\ln(1+e^x)$] also perform comparably. 
Linear or piecewise linear functions such as ReLU or triangular function cannot reproduce the L63 dynamics within the tested range of training data and neurons. 
We shall use a geometric perspective to show that such failure may be due to the inability of (piecewise) linear functions to develop the nonlinear flow required by chaos. 

\begin{table}[ht]
\caption{Parameters of the 4-neuron NN trained on 40 data points from L63. Matrices $\mathbf{W}_1$, $\mathbf{W}_2$, $\mathbf{b}_1$, $\mathbf{b}_2$ are as in \eqref{eq:NN}. $\mathbf{S}_{jj}$ are both the elements of diagonal matrix $\mathbf{S}$ and the singular values of $\mathbf{W^*}$.}\label{tab:parameters1}
\centering
\ra{1.2}
\begin{tabular}{c|c}
\toprule
Matrix			& Values 					\\
\hline
$\mathbf{W}_1$		&$ \begin{array}{rrrr} 	0.0091	&0.0008 	&-0.0004	\\ 
										0.0140	&0.0063	&-0.0016	\\
										0.0061	&0.0023	&-0.0049	\\
										0.0085	&0.0036	&0.0041	\end{array}$\\
\hline
$\mathbf{b}_1^{\mathrm{T}}$		&$ \begin{array}{rrrr}	0.1697   &-0.6054   &-0.0449   &0.1773	\end{array}$\\
\hline
$\mathbf{W}_2$		&$ \begin{array}{rrrr} 	94.6004    	&8.7248   	&-8.0364	&3.0535	\\
  										-349.8684  	&11.3885	&207.0634&227.4161	\\
 										32.1244	&93.9784	&-214.6608	&11.9787		\end{array}$\\
\hline
$\mathbf{b}_2^{\mathrm{T}}$		&$ \begin{array}{rrrr}	-12.1241   &34.2950   &33.6097	\end{array}$\\
\hline
$\mathbf{S}_{jj}$		&$ \begin{array}{rrrr}	2.7988 	&1.2134	&0.6438	&0.0000	\end{array}$\\
\bottomrule
\end{tabular}
\end{table}

Our experiments show that NN learns L63 dynamics efficiently with a small number of data and neurons.
We now analyze a 4-neuron NN, the smallest viable network.  Trained on only 40 data points, it  accurately reconstructs the strange attractor. 
Table\;\ref{tab:parameters1} shows the network's parameters after training. 
The blue solid trajectory in Fig.\;\ref{fig:NN_predictions_small_num} follows the L63 flow, 
and the red dashed one follows the flow of the trained NN. 
They interlace with each other, and both trace out the well-known Lorenz attractor. 
NN trajectories starting from other locations on the attractor follow the same behavior and do not diverge. 
The close resemblance between the two structures indicates that the NN's dynamics is similar to L63. 
This is especially notable since this NN is very simple with only 4 neurons and is trained on 40 data points; far fewer than the typical training practice in the literature. 
The root-mean-square (RMS) prediction error on testing data 
%(Fig.\;\ref{fig:errors}a) 
(Fig.\;S2a) 
decreases with increasing number of neurons and training data, consistent with Zhang (2017)~\cite{Zhang2017b}. We will not focus on the prediction error since it has already been discussed in past work. Instead, we compare the short-term and long-term predictability of the two systems in the following section.

\subsection{Comparison of predictability}
\label{sec:FTLE}

To quantify the predictability of NN and L63 in terms of local divergence rates, we use maximum finite-time Lyapunov exponent (FTLE)\;{\cite{Haller2001}}. 
Here, maximum FTLE describes the largest possible exponential divergence rate of nearby trajectories originating from the L63 attractor: 
\begin{equation}
    \lambda_{\text{max}} := \frac{1}{N_t}\ln\frac{\left|\max\limits_{\delta\mathbf{x}_0}\delta\mathbf{x}_{N_t}\right|}{|\delta\mathbf{x}_0|} = \frac{1}{N_t}\ln{\sqrt{\sigma_{\text{max}}}},
    \label{eq:FTLE_definition}
\end{equation}
where $\delta\mathbf{x}_0$ is the initial perturbation that achieves maximum divergence, $\delta\mathbf{x}_{N_t}$ is the distance after $N_t$ steps, and $\lambda_{\text{max}}$ is the maximum FTLE. FTLE reduces to the Lyapunov exponent\;{\cite{Barreira2002}} when $N_t\rightarrow\infty$ and $\delta\mathbf{x}_0\rightarrow0$. In Eq.\;\eqref{eq:FTLE_definition}, $\lambda_{\text{max}}$ is calculated using the largest eigenvalue ($\sigma_{\text{max}}$) of $\mathbf{J}_{N_t}^\mathrm{T}\mathbf{J}_{N_t}$, where $\mathbf{J}_{N_t}$ is the Jacobian matrix evaluated using perturbations around $\mathbf{x}_0$ (see supplementary text 
%\ref{sec:FTLE_cal}).
S1). 

\begin{figure}[ht]
\centering
\includegraphics[width=0.8\linewidth]{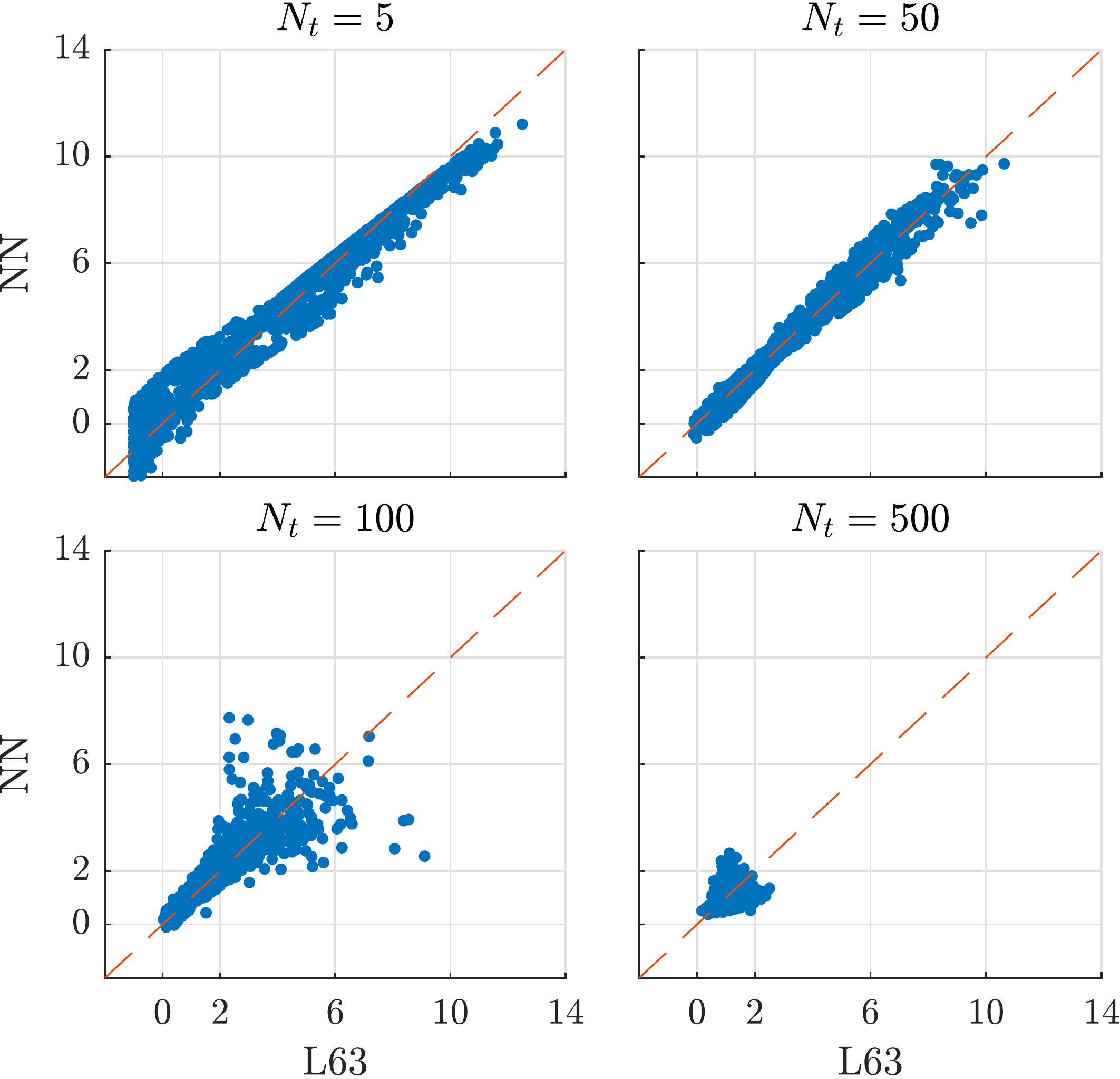}
\caption{One-to-one scatter plot of FTLE with L63 ($x$ axis) and NN ($y$ axis). The NN used in this plot is the same as that in Fig.~{\ref{fig:NN_predictions_small_num}}. The panels (from left to right, top to bottom) correspond to increasing integration steps, $N_t$. }
\label{fig:FTLE}
\end{figure}

We now compare the FTLE of NN and L63 dynamics on the attractor. 
Because FTLE depends on trajectories' initial locations, the starting points of L63 and NN trajectories should be close for a valid comparison. 
For the selected NN with 4 neurons and trained with 40 data points, we first generate points using the same generation process as in section \ref{sec:compactNN} to represent the NN attractor ($\mathcal{A}_{\text{NN}}$). 
Second, we randomly choose 2000 starting positions on $\mathcal{A}_{\text{L63}}$. 
Each point is paired with the closest point on $\mathcal{A}_{\text{NN}}$. 
A pair of trajectories then initializes from each pair of points, 
and the former trajectory follows the L63 flow, whereas the latter follows the NN flow. 
The FTLE of the trajectory pairs is compared under different integration steps: $N_t = 5, 50, 100, 500$ as shown in Fig.\;{\ref{fig:FTLE}}. 
When $N_t = 5, 50$, NN accurately reproduces local divergence rates over the whole attractor, showing that the two systems' short-term predictability agrees with each other. 
As $N_t$ increases ($N_t$ = 100), the chaotic nature of the two flows begin to emerge so that some NN-L63 trajectory pairs end up being faraway from each other. 
For the trajectory pairs whose end points are separated by a large distance, the FTLE measures the accumulated divergence rates along distinct regions in the phase space. Therefore, we expect the FTLE correspondence to diverge. 
At $N_t$ = 500, the FTLE collapses to the long-term maximum Lyapunov exponent of L63 (roughly 0.91 as in\;{\cite{Viswanath1998}}), indicating that the two systems' long-term behavior is also similar. 

\begin{figure}[ht]
\centering
\includegraphics[width=0.8\linewidth]{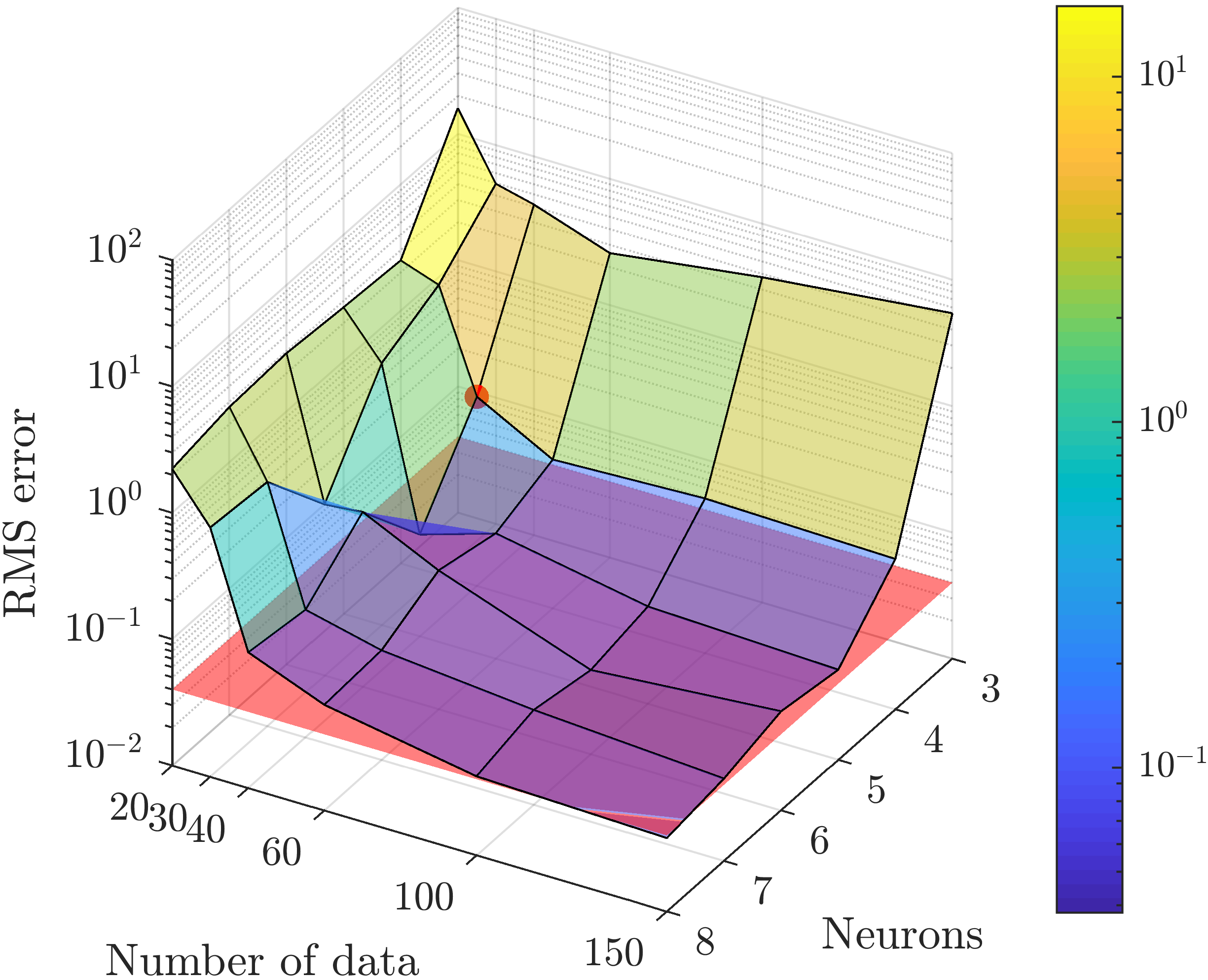}
\caption{The FTLE test errors (also root-mean-square) of neural networks for each neuron and number-of-data configuration. The FTLE is calculated with $N_t = 50$ and averaged over 2000 trajectories that are randomly initialized on the attractor. The red dot represents the example configuration used in Figs.\;{\ref{fig:NN_predictions_small_num}} and {\ref{fig:FTLE}}. The red surface is located at $z=0.04$. }
\label{fig:FTLE_errors}
\end{figure}

The FTLE errors at $N_t$ = 50 generally decrease with increasing numbers of neurons and training data points (Fig.\;{\ref{fig:FTLE_errors}}). This trend is similar to the decreasing trend of the RMS prediction errors 
%(Fig.\;\ref{fig:errors}). 
(Fig.\;S2a). 
The reduction in the FTLE and prediction errors follows from the bias and variance trade-off\;{\cite{Goodfellow2016}}: increased complexity in learning models generally translates into lower bias in prediction, provided that regularization techniques prevent the learning algorithm from entering the high-variance (overfitting) regime. 
Although the learning process only minimizes prediction error, it also improves the agreement in FTLE. 

\begin{figure}[ht]
    \centering
    \includegraphics[width=0.8\linewidth]{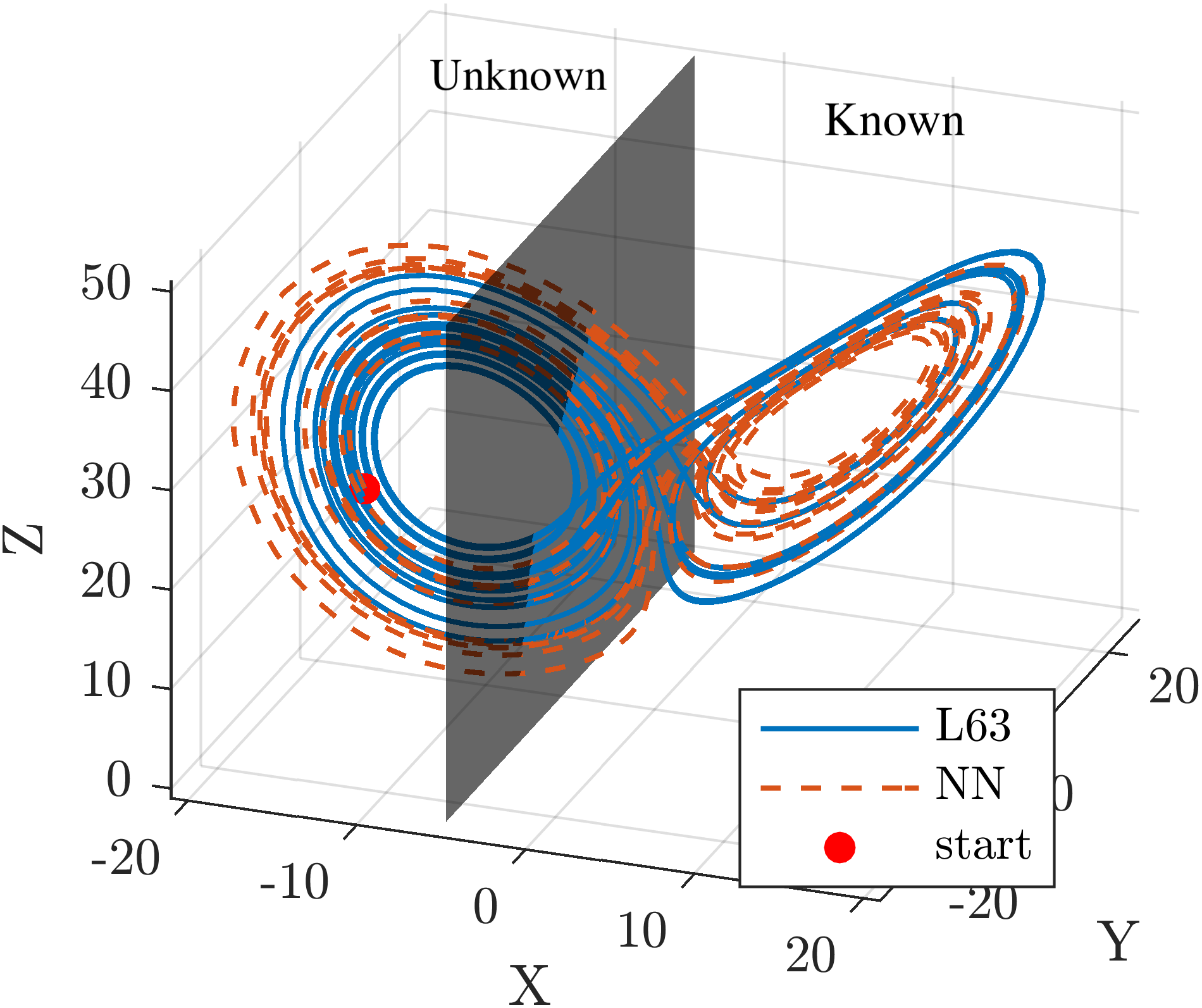}
\caption{Similar to Fig. {\ref{fig:NN_predictions_small_num}}, but the red-dashed trajectory is produced by a 5-neuron NN trained on 100 data points sampled from the $X>-5$ part of the L63 attractor. The region to the right of the grey partition is the training data range, and the region to the left is unknown to the NN. }
\label{fig:NN_predictions}
\end{figure}

Remarkably, NN can extrapolate with incomplete training data from a segment of the attractor. 
Similar to Fig.\;{\ref{fig:NN_predictions_small_num}}, Fig.\;{\ref{fig:NN_predictions}} compares two same-origin trajectories separately predicted by NN and L63. 
In this case, the NN has 5 neurons and is trained on 100 data points sampled from the $X>-5$ part of $\mathcal{A}_{\text{L63}}$, 
which amounts to knowing 73\% of the attractor structure. 
Despite originating from the unknown region of $X\le-5$, the NN trajectory still traces a smooth path which closely resembles the original attractor in the extrapolated region. 
The NN trajectory is close to the L63 one in the first 100 timesteps and then bifurcate onto the two branches of the attractor (not shown). 
The one-to-one correspondence of FTLE between L63 and the 5-neuron NN trained on the incomplete data is similar to Fig.\;{\ref{fig:FTLE}} 
%(see Fig.\;\ref{fig:FTLE_half_domain}). 
(see Fig. S3). 

\section{A geometric perspective of the NN flow}
\label{sec:geo}

The agreement in FTLE between NN and L63 and NN's extrapolation skill suggest that NN has surprisingly good efficacy in learning chaotic dynamics, which UAP cannot explain. 
The UAP states that NN can approximate maps like $\mathbf{\Phi}_{\text{L63}}$ arbitrarily well, but this does not explain NN's efficacy in reconstructing the strange attractor nor its extrapolation skill. 
We instead draw inspiration from the exact mathematical correspondence between the geometric Lorenz flow and L63\;{\cite{Guckenheimer1979, Tucker2002}} 
(see supplementary text 
%\ref{sec:geometric})
S2)
and provide a geometric understanding of why NN can emulate chaos efficiently. 

\subsection{Mathematical formulation}

The dynamics of NN (Eq.\;\ref{eq:NN}) can be seen as a map in a multi-dimensional Riemann space (this interpretation was also used in classification problems\;{\cite{Hauser2017}}). In the discrete map of the 4-neuron NN in the previous section (table \ref{tab:parameters1}), the input vector $\mathbf{x}$ in the 3D phase space is mapped into a 4-D \textit{neuron space} and then mapped back to the phase space. Let an $N_t$-step phase-space trajectory be $\mathcal{L}_0^{N_t} = \{\mathbf{x}_0, \mathbf{x}_1, ..., \mathbf{x}_{N_t}\}$, $N_t \geq 2$. From step $n$ to $n+1$, there exists a 4-D intermediate vector $\mathbf{y}$ in the neuron space: 
\begin{equation}
\mathbf{y}_{n + 1} = g(\mathbf{W}_1\mathbf{x}_n + \mathbf{b}_1),~~~ (n = 0, 1, ..., N_t - 1).
\label{eq:geometric1}
\end{equation}
We refer to $\mathbf{y}$ as the \textit{neuron vector}. The recurrence relation of $\mathbf{y}$ is 
\begin{equation}
\mathbf{y}_{n + 1} = g(\mathbf{W^*}\mathbf{y}_n + \mathbf{b^*}), ~~~ (n = 1, 2, ..., N_t - 1).
\label{eq:geometric2}
\end{equation}
where $\mathbf{W^*} = \mathbf{W}_1\mathbf{W}_2$ is a 4-by-4 matrix, 
and $\mathbf{b^*} = \mathbf{W}_1\mathbf{b}_2 + \mathbf{b}_1$ is a 4-by-1 vector. 
We denote Eq.\;\eqref{eq:geometric2} as the \textit{neuron map}. Understanding the neuron map is equivalent to understanding the dynamics of NN because the neuron map {\eqref{eq:geometric2}} is only different from the NN map {\eqref{eq:NN}} by a homomorphism, Eq.\;\eqref{eq:geometric1}. 

The neuron map comprises 4 sub-steps: rotation, stretch, rotation, and compression. 
To see how, we use singular value decomposition (SVD) to rewrite $\mathbf{W^*}$ as $\mathbf{U}\mathbf{S}\mathbf{V}^{\mathrm{T}}$, and rewrite Eq.\;\eqref{eq:geometric2} as 
\begin{equation}
\mathbf{y}_{n + 1} = g(\mathbf{U}\mathbf{S}\mathbf{V}^{\mathrm{T}}\mathbf{y}_n + \mathbf{b^*}), 
\label{eq:geometric3}
\end{equation}
where $\mathbf{U}$ and $\mathbf{V}$ are 4-D orthonormal matrices, and $\mathbf{S}$ is a diagonal matrix of rank 3\footnote{Because $\mathbf{W^*}$ is the product of a 4-by-3 matrix and a 3-by-4 matrix, the rank of $\mathbf{W^*}$ is at most 3, and so is the rank of $\mathbf{S}$. For the 4-neuron network at question, $\mathbf{S}$ has 3 diagonal elements.}. This expression suggests that every neuron vector ($\mathbf{y}$) is sequentially rotated by $\mathbf{V}^{\mathrm{T}}$, stretched by $\mathbf{S}$, rotated by $\mathbf{U}$, and compressed by $g(\cdot)$. 
Note that ``rotation" here takes the generalized sense of orthogonal transformation while preserving the $L^2$ norm, and this also includes reflection. 
The sigmoidal function only applies a compressing effect because it squashes the distance between any two points on the real line.

The compressing and stretching in the NN map are seen more clearly through the growth of perturbations in the neuron space. Let $\delta\mathbf{y}$ be a small perturbation between two trajectories near location $\mathbf{y}$. Linear expansion of Eq.\;\eqref{eq:geometric2} gives the perturbation at the next timestep
\begin{equation}
\delta \mathbf{y}' = g'\left(\mathbf{W}^*\mathbf{y} + \mathbf{b}^*\right)\odot\mathbf{W}^*\delta\mathbf{y},
\label{eq:perturbation2}
\end{equation}
where $\odot$ denotes element-wise (or Hadamard) product. Let $G_{jj} = g'\left(\sum_{i = 1}^L{W}^*_{ji}y_i + b^*_j\right)$ and $\mathbf{G} = \mathrm{diag}\{G_{11}, G_{22}, ...\}$, and rewrite $\mathbf{W}^*$ with its singular value decomposition, the squared distance at the next timestep can be written as
\begin{equation}
|\delta \mathbf{y}'|^2 = |\mathbf{G}\mathbf{U}\mathbf{S}\mathbf{V}^{\mathrm{T}}\delta\mathbf{y}|^2. 
\label{eq:perturbation3}
\end{equation}
In Eq.\;\eqref{eq:perturbation3}, all elements in $\mathbf{G}$ are smaller than 1 because $g'(x) \in (0, 1],\; \forall x\in\mathbb{R}$. 
This implies that $\mathbf{G}$, or the activation function, only applies a compressing effect on perturbations, 
consistent with our analysis of Eq.\;\eqref{eq:geometric3} above. 
Consequently, at least one element in $\mathbf{S}$ must be larger than 1 to obtain one or more unstable directions as required by chaos\;{\cite{Devaney1989}}. 
For the 4-neuron NN in table\;\ref{tab:parameters1}, $\mathbf{S}$ has two elements larger than 1 and therefore applies stretching in two dimensions in the neuron space. 
Given the information of $\mathbf{y}$, $\mathbf{G}$ controls the degrees of compression in each dimension of the neuron space. 
The rotations by $\mathbf{U}$ and $\mathbf{V}^{\mathrm{T}}$ control the orientations of the compression and stretching by $\mathbf{G}$ and $\mathbf{S}$ while not changing the magnitude of the perturbations.  

The above framework can be easily generalized into an $N$-hidden-layer network. The neuron vector dynamics is ambiguous here as multiple hidden layers have multiple neuron vectors. We can nevertheless apply the same method to perturbations in the phase space. For a perturbation of $\delta\mathbf{x}$ around $\mathbf{x}$, its squared length at the next timestep is
\begin{equation}
|\delta \mathbf{x}'|^2 = |\mathbf{W}_{N+1}\mathbf{G}_N\mathbf{W}_N...\mathbf{G}_1\mathbf{W}_1\delta\mathbf{x}|^2. 
\label{eq:perturbation5}
\end{equation} 
Gradient matrix $\mathbf{G}_i = \mathrm{diag}\{g'(\mathbf{W}_i\mathbf{y}_{i-1} + \mathbf{b}_i)\}$, in which $\mathbf{y}_{i-1}$ is the neuron vector of the $i$th layer for $i>1$, and $\mathbf{y}_0 = \mathbf{x}$. $\mathbf{W}_i$ is the weight matrix that connects layers $i$ and $i+1$. The weight and gradient matrices consecutively parameterize multiple stretching and compressing operations in a single NN map. 

\subsection{Topological mixing in NN with the H\'enon map}

The stretch and compression sub-steps in neuron maps are thought of as the typical way to give rise to topological mixing and chaos (although strictly speaking, it is neither the necessary nor the sufficient condition\;{\cite{Ruelle2006}}). The ability to obtain these geometric operations makes NN very good at approximating discrete chaotic maps. Since it is challenging to visualize the 4-D neuron-space dynamics in the NN trained on the L63 system, we now use a 2-neuron NN trained on the H\'enon map for illustration. The H\'enon map is a discrete 2D chaotic map designed such that trajectories in the $x$-$y$ plane are stretched in one direction and compressed in the other\;{\cite{Henon1976}}. The map comprises three sub-steps: an area-preserving stretch, a compression, and a reflection along $x = y$:
\begin{equation}
\begin{aligned}
(x_1, y_1) &= (x, 1-ax^2 + y), &\text{(stretch)}\\
(x_2, y_2) &= (bx_1, y_1), &\text{(compression)}\\
(x', y') &= (y_2, x_2), &\text{(reflection)}
\label{eq:henon}
\end{aligned}
\end{equation}
where $(x, y)$ is the starting location and $(x', y')$ is the finishing location after one iteration. 
We set $a = 1.4$, $b = 0.3$ and generate training data.
A 2-neuron network is trained with only 20 randomly-sampled data points following the same training procedure as L63. 
The reconstructed strange attractor is virtually indistinguishable from the original one
%(Fig.\;\ref{fig:henon}). 
(Fig.\;S4). 
Table {\ref{tab:parameters2}} shows the parameters of this network after training. 

\begin{table}[!ht]
\caption{Parameters of the 2-neuron NN trained with 20 data points of the H\'enon map. }\label{tab:parameters2}
\centering
\begin{tabular}{c|c|c|c}
\toprule
Matrix			& Values 					&Bias		&Values \\
\hline
$\mathbf{W}_1$	
&$\begin{array}{rr} 
        0.0960 & 0.0043  \\ 
        -0.0866 & 0.0041 
\end{array}$
&$\mathbf{b}_1^{\mathrm{T}}$	&$0.8688, 0.9188$\\
\hline
$\mathbf{W}_2$	
&$\begin{array}{rr} 
    220.7978 & 263.0327  \\ 
    3.0292 & -3.6975
\end{array}$
&$\mathbf{b}_2^{\mathrm{T}}$	&$-344.5050, 0.5593$\\
\bottomrule
\end{tabular}
\end{table}

\begin{figure*}[ht]
\centering
\includegraphics[width=0.8\linewidth]{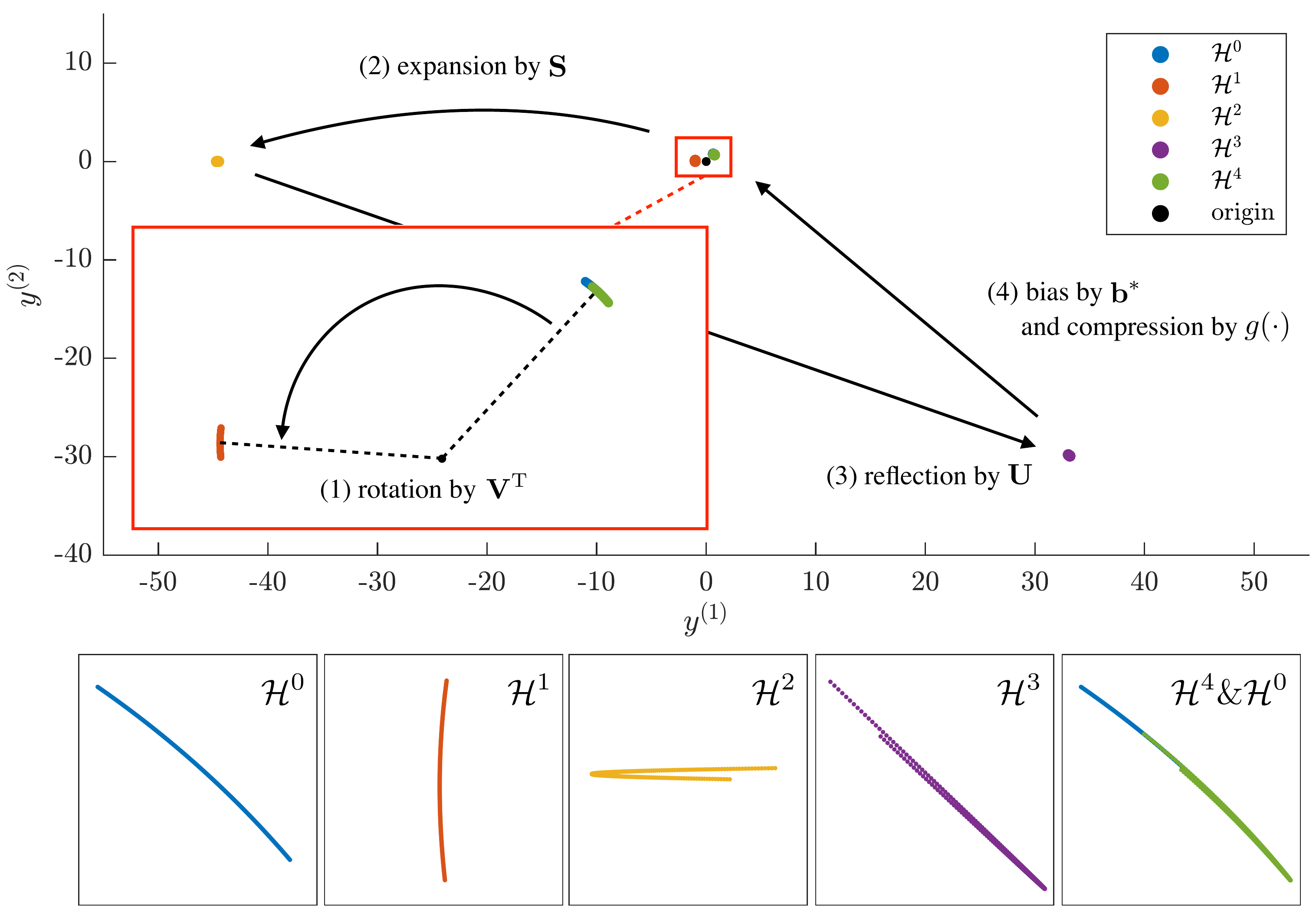}
\caption{Schematic of an iteration of the neuron map described by the 2-neuron NN trained on the H\'enon map. The positions of points in the neuron space at each sub-step is shown in the upper figure, and the detailed structure is sketched in the lower panels. $\mathcal H^0$ (blue) is rotated counter-clockwise to $\mathcal H^1$ (orange), stretched and contracted to $\mathcal H^2$ (yellow), reflected to $\mathcal H^3$ (purple), and then compressed to $\mathcal H^4$ (green) which occupies the same region as $\mathcal H^0$. The first step is magnified in the inset. }
\label{fig:mapping}
\end{figure*}

The 2-neuron network shows how the stretching, rotation, and compression operations take place in the neuron map. 
Let $\mathcal{H}$ be a group of points that form a straight line in the phase space. 
$\mathcal{H}$ is initialized into the neuron space by {\eqref{eq:geometric1}} as $\mathcal{H}^0$.
$\mathcal{H}^0$ is shown as the blue dots in Fig.\;{\ref{fig:mapping}}, and the two neuron-space dimensions are denoted as $y^{(1)}$ and $y^{(2)}$. 
Then $\mathcal{H}^0$ undergoes a series of geometric maps following the NN flow. 
First, $\mathbf{V}^{\mathrm{T}}$ rotates $\mathcal{H}^0$ around the origin by $130.0^\circ$ counter-clockwise to $\mathcal{H}^1$ (red). 
Next, $\mathbf{S}$ stretches $\mathcal{H}^1$ in $y^{(1)}$ and compresses it in $y^{(2)}$, yielding $\mathcal{H}^2$ (yellow). $\mathcal{H}^2$ is reflected by $\mathbf{U}$ along a line of $69.0^\circ$ to the positive $y^{(1)}$ axis and becomes $\mathcal{H}^3$ (purple). The addition of bias $\mathbf{b}^*$ and element-wise compression by $g(\cdot)$ transform $\mathcal{H}^3$ to $\mathcal{H}^4$ (green). The next step of the neuron map then initiates from $\mathcal{H}^4$. Compared with $\mathcal{H}^0$, $\mathcal{H}^4$ extends along the point cloud's principal direction and wraps around the lower-right tip of $\mathcal{H}^0$. Hence, each iteration of the neuron map is effectively a horseshoe transformation that leads to topological mixing and chaos. 

The geometric view also explains why single-layer networks with linear and piecewise linear activation functions 
such as ReLU fail to reproduce the L63 dynamics within the tested range of neurons and training data. 
For linear functions, or in the linear regime of piecewise linear activation functions, $\mathbf{G}$ in Eq.\;\eqref{eq:perturbation3} is a constant matrix. 
Hence, they cannot emulate chaos because perturbations in each dimension either expands or contracts indefinitely, yielding a fixed point if all eigenvalues of $\mathbf{G}\mathbf{W}^*$ are smaller than 1, or a diverging system if otherwise. 
Indeed, linear models dramatically fail when used to model chaotic dynamics\;\cite{Dudul2005}. 
NNs with nonlinear activation functions are trained to use $\mathbf{G}$ as a function of $\mathbf{y}$ to modify and control the degree of compression in neuron space dimension, thereby making NN an effective tool to emulate chaotic dynamics. 
However, this does rule out the possibility that some piecewise linear functions can produce chaotic behaviors. 
For example, the piecewise-linear triangular basis function expresses a folding operation of $x\in(-1, 1)$ onto $x'\in(0, 1)$. Therefore, if a stretch ratio of 2 and a bias of -1 are applied, it's conceivable to use this activation function to reproduce a horseshoe map with only one hidden unit. 

%The example of the H\'enon map shed light on the geometric connection between NN and chaotic maps, and it also hints at why NN is able to extrapolate outside the training data range. The training data feeds part of the behavior for the left branch by providing information for the $-5<X<0$ of the attractor

\section{Lower-bounding the number of neurons}
\label{sec:bounds}

The necessary numbers of neurons that reproduce the L63 and H\'enon maps are surprisingly small compared to predictions of previous theoretical results. 
Since the Euler-forward scheme of {\eqref{eq:L63}} is a 3D ($n$=3) polynomial with a degree of at most $d$=2, 
we use previous theoretical results on learning polynomials with NNs\;{\cite{Barron1993, Andoni2014}} to establish lower bounds on the necessary number of neurons. 
We assume that polynomials characterize the true dynamics, but the learning system doesn't know the exact coefficients for each term. 
The number of neurons ($L$) for learning a polynomial with RMS error target $\epsilon$ is bounded by $L = \Omega(n^{6d}/\epsilon^3)$ according to\;{\cite{Andoni2014}}. 
This is a rather rough estimate as more than $5\times10^5$ nodes are needed when $\epsilon \sim 1$ (for H\'enon map, the estimate is $4\times10^3$).

Matching the near-equilibrium norms of neural and polynomial regression\;{\cite{Trautner2021}} gives a more reasonable bound. A PolyNet\;{\cite{Trautner2019}} asymptotically needs $L={n+d\choose d}-(n+1)\approx6$ hidden nodes to exactly match a full polynomial $(n$=$3,d$=$2)$.  
An incomplete polynomial with several coefficients fixed a priori can correspond to the L63 system exactly, and only two hidden nodes are needed for the regression norms to match. However, this is an unreasonable amount of knowledge for bottom-up learning to assume. 
% L={n+d\choose d}-(n+1)\sim6 is because there're n(n+d\choose d) parameters for the polynomial learner, and $2nL+L+n+n^2$ for a FNN learner with $n^2$ linear units and $n$ constant bias units that bypasses the nodes. Equating the two gives $2nL+L+n+n^2 \sim n(n+d\choose d)$. 
The standard network (NN) used here {\eqref{eq:NN}} has an asymptotic bound of $L\sim\frac{n}{2n+1}\left[{n+d\choose d} - 1 \right]\approx 5$ neurons. 
% $2nL+L+n \sim n(n+d\choose d)$ gives the above result
Note that these estimates based on the training process do not provide a training error guarantee. 

A more direct but less rigorous bound can be obtained via a Taylor-expansion of the sigmoid function to the third order: $\tanh(x) = x - x^3/3 + O(x^5)$, which allows \eqref{eq:NN} to be modeled as a polynomial of degree 3 (NN polynomial). We further require all coefficients of the NN polynomial to be equal to those in {\eqref{eq:L63}}. Then for an NN with $L$ hidden nodes, biases, and $n$-dimensional input/output, a total of $2nL+n+L$ parameters should satisfy $3 {n+3 \choose 3}$ constraining equations. The parameters in NN should be under-determined for a good fit, i.e., $2nL+n+L \ge 3{n+3 \choose 3}$. Hence, at least $L = \ceil{(3{n+3 \choose 3}-n)/(2n+1)} = 9$ hidden nodes are needed. To obtain an error estimate, we substitute table {\ref{tab:parameters1}} into the NN polynomial to obtain $\mathbf{\Phi}_{\mathrm{NN-poly}}$ and calculate the expected error over data sampled from the $\text{L63}$ attractor: $\epsilon^2 = \langle(\mathbf{\Phi}_{\mathrm{NN-poly}}-\mathbf{\Phi}_{\text{L63}})^2\rangle_{\mathcal{A}_{\text{L63}}}$. Five thousand random samples give a normalized error of $\epsilon \sim 0.14$. Therefore, this estimation gives a lower bound of 9 neurons at the error level of at most $0.14$. Thus, the network sizing in the experiments cannot be interpreted as overfitting.

\section{Conclusion and discussion} 
\label{sec:concl}

We have demonstrated that single-hidden layer feedforward neural networks with nonlinear activation functions can emulate chaotic systems such as Lorenz-63 and H\'enon map with surprising efficacy. 
Our results suggest that NN is potentially a good candidate to represent a broad class of chaotic dynamics. It learns  efficaciously from data and offers good generalization skill. 
Such success is explained by revealing NN's structural similarity to the chaotic maps in terms of the stretching and compression operations. 
The high-dimensional rotations are also important to reproduce the flow-like dynamics of L63. 
Therefore, NN may serve as a suitable non-parametric model for chaotic systems in data-driven problems 
because, contrary to conventional thinking, it requires low complexity in model design and is not data-hungry. 
Conversely, one could also consider the trained NNs as a unifying formulation of dissipative chaotic systems because NN reproduces the H\'enon and L63 maps under the same mathematical framework. 

On the other hand, the compression operation imposed by the sigmoid function makes NN of the form of {\eqref{eq:NN}} preferable to emulate low-dimensional dissipative systems. Its ability to model chaotic non-dissipative Hamiltonian dynamics and systems of much higher dimensionality is yet to be tested. 
More work is also needed, possibly with the aid of Riemann geometry, to fundamentally understand the geometric operations in the high-dimensional neuron space.

% or


\iffalse
\appendices
\section{Proof of the First Zonklar Equation}
Appendix one text goes here.

\section{}
Appendix two text goes here.
\fi

\section*{Acknowledgment}

Sai Ravela advised Ziwei Li. Support from ONR grant N00014-19-1-2273, NSF award 1749986, the MIT Environmental Solutions Initiative, and the John S. and Maryann Montrym Fund are gratefully acknowledged.

\ifCLASSOPTIONcaptionsoff
  \newpage
\fi



%\bibliographystyle{IEEEtran}

\bibliographystyle{./IEEEtran}
\bibliography{./NN_Lorenz}
%\bibliography{/Users/ziweili/Documents/references/library.bib}


\begin{IEEEbiography}[{\includegraphics[width=1in,height=1.25in,clip,keepaspectratio]{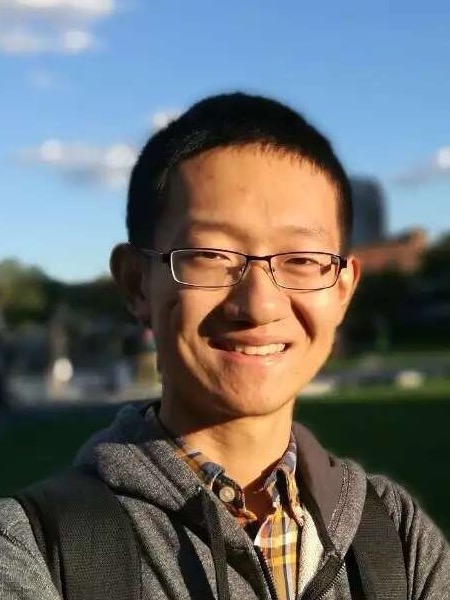}}]{Ziwei Li}
is a Ph.D. candidate in the Department of Earth, Atmospheric and Planetary Sciences 
at the Massachusetts Institute of Technology. 
He is interested in understanding atmosphere dynamics using simple physical and stochastic models, 
as well as applying machine-learning algorithms to scientific problems in a physics-informed way. 
Ziwei Li received his Bachelor of Science degree from Peking University in 2016. 
\end{IEEEbiography}

\begin{IEEEbiography}[{\includegraphics[width=1in,height=1.25in,clip,keepaspectratio]{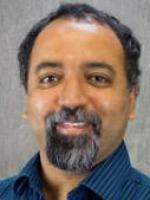}}]{Sai Ravela}
directs the Earth Signals
and Systems Group (ESSG) at the Massachusetts Institute of Technology. 
His primary research interests are dynamic data-driven  modeling, observation, and optimization of  
stochastic systems, including learning systems, for earth, planet, climate and life applications. Dr. Ravela holds a Ph.D. in Computer Science from
the University of Massachusetts at Amherst (2003). 
He is a recipient of the 2016 MIT Infinite Kilometer Award for research and mentoring excellence. He is the CTO of Windrisktech LLC.
\end{IEEEbiography}

%\newpage


%\vfill


\end{document}

\pagestyle{plain}
\setcounter{page}{1}

\maketitle

\begin{abstract}
This file contains the supplementary figures and texts for article "Neural Networks as Geometric Chaotic Maps". We describe the numerical computation of FTLE in section \ref{sec:FTLE_cal} and the connection between the geometric Lorenz map and the classic ODE Lorenz map in section \ref{sec:geometric}, and then lay out supplementary figures in \ref{sec:supp_figures}. 
\end{abstract}

\clearpage

\section{Numerical computation of FTLE}
\label{sec:FTLE_cal}

We define the map over $N_t$ time steps as $\Phi_{0\rightarrow N_t}: \mathbf{x}_0\mapsto\mathbf{x}_{N_t}$. When a perturbation $\delta\mathbf{x}$ around $\mathbf{x}_0$ is sufficiently small and $N_t$ is finite, the resulting distance, $\delta\mathbf{x}_{N_t}$, can be linearly approximated as 
\begin{equation}
    \delta\mathbf{x}_{N_t} = \mathbf{J}_{N_t}(\mathbf{x}_0)\delta\mathbf{x} + O(|\delta\mathbf{x}|^2), 
    \label{eq:FTLE1}
\end{equation}
%$\left.\frac{\partial\Phi_{0\rightarrow T}(\mathbf{x})}{\partial \mathbf{x}}\right|_{\mathbf{x}_0}$
where $\mathbf{J}_{N_t}(\mathbf{x}_0)$ is the Jacobian of mapping $\Phi_{0\rightarrow {N_t}}$ evaluated by forward-propagating perturbations around $\mathbf{x}_0$ for $N_t$ steps. The magnitude of $\delta\mathbf{x}$ is $10^{-9}$ for all 6 directions in the 3D phase space. Neglecting higher-order terms in Eq.\;\eqref{eq:FTLE1} and taking its norm, we have
\begin{equation}
    |\delta\mathbf{x}_{N_t}|^2 = \delta\mathbf{x}^\mathrm{T}\mathbf{J}_{N_t}(\mathbf{x}_0)^\mathrm{T}\mathbf{J}_{N_t}(\mathbf{x}_0)\delta\mathbf{x},
    \label{eq:FTLE2}
\end{equation}
Then, the problem of finding the direction of $\mathbf{x}_0$ that maximizes perturbation growth rate reduces to solving for the eigenvector that corresponds to the largest eigenvalue of matrix $\mathbf{J}_{N_t}(\mathbf{x}_0)^\mathrm{T}\mathbf{J}_{N_t}(\mathbf{x}_0)$, and the largest growth rate corresponds to the largest eigenvalue of the matrix. The maximum FTLE is therefore evaluated as
\begin{equation}
    \lambda_{\text{max}} = \frac{1}{N_t}\ln{\sqrt{\sigma_{\text{max}}}}, 
    \label{eq:FTLE3}
\end{equation}
where $\sigma_{\mathrm{max}}$ is the largest eigenvalue of $\mathbf{J}_{N_t}(\mathbf{x}_0)^\mathrm{T}\mathbf{J}_{N_t}(\mathbf{x}_0)$. Eq.\;\eqref{eq:FTLE3} is used to find the local maximum FTLE in section 
%\ref{sec:FTLE}. 
II-B.

\section{Exact correspondence between the geometric Lorenz map and L63}
\label{sec:geometric}

\begin{figure}[ht]
\centering
\includegraphics[width=0.855\textwidth]{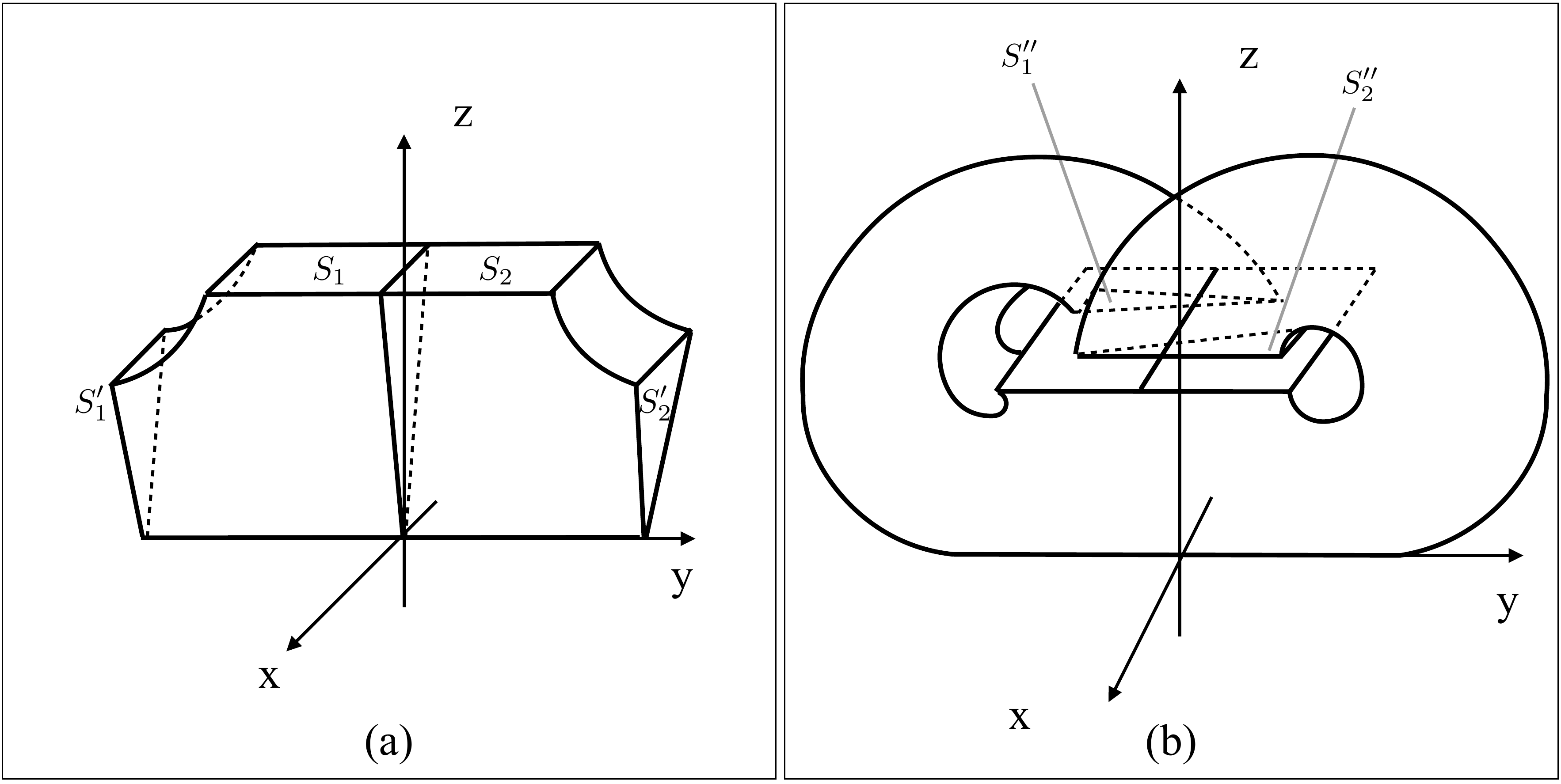}
\caption{\onehalfspacing Schematic of the geometric Lorenz flow, similar to Figs.\;1 and 2 in {\cite{Guckenheimer1979}}. (a) A rectangle, $S$, parallel to the $\mathrm{x}$-$\mathrm{y}$ plane intercepting the positive $\mathrm{z}$ axis is divided into $S_1$ ($y<0$) and $S_2$ ($y>0$). $S_1$ is moved to the lower left, compressed in $\mathrm{x}$ and becomes triangle $S_1'$, whereas $S_2$ is moved symmetrically to the right and becomes $S_2'$. (b) $S_1'$ and $S_2'$ are swirled and mapped back onto the original rectangle as $S_1''$ and $S_2''$, respectively. $S_1''$ satisfies $x<0$, and $S_2''$ satisfies $x>0$, so that they occupy mutually exclusive regions. After multiple iterations of this geometric flow, two fractal attractors emerge on the joint rectangle of $S_1$ and $S_2$. Note that the directions of x and y are rotated 45$^\circ$ counter-clockwise compared to the original L63 system for ease of illustration.}
\label{fig:lorenz_geometric}
\end{figure}

The fact that L63 system is a first-order ordinary differential equation set and cannot be analytically solved means an exact solution can never be obtained. Fortunately, Ref.\;{\cite{Tucker2002}} rigorously proved that the numerical solution of the original dynamical equations has the same topological properties as the \textit{geometric Lorenz flow} proposed by {\cite{Guckenheimer1979}}, which has been extensively studied since its first publication. The geometric flow has a compressing operation in the $\mathrm{x}$ direction and a stretching mainly in the $\mathrm{y}$-$\mathrm{z}$ plane (Fig.\;{\ref{fig:lorenz_geometric}}). The stretching has two important properties: first, it has an anti-symmetric $\mathrm{x}$ component such that the surfaces $S_1$ and $S_2$, which were originally separated in the $\mathrm{y}$ direction are now separated in $\mathrm{x}$; second, it stretches in the $\mathrm{y}$ direction such that $S_1''$ and $S_2''$ becomes \textit{mixed} on the original $S_1$ and $S_2$ manifold. These geometric operations are such that topological mixing takes place on the joint surface of $S_1$ and $S_2$, leading to chaotic behavior.

\bibliography{./NN_Lorenz}
\bibliographystyle{ieeetr}

\newpage

\section{Supplementary figures}
\label{sec:supp_figures}

\begin{figure}[ht!]
\centering
\includegraphics[width=1.0\textwidth]{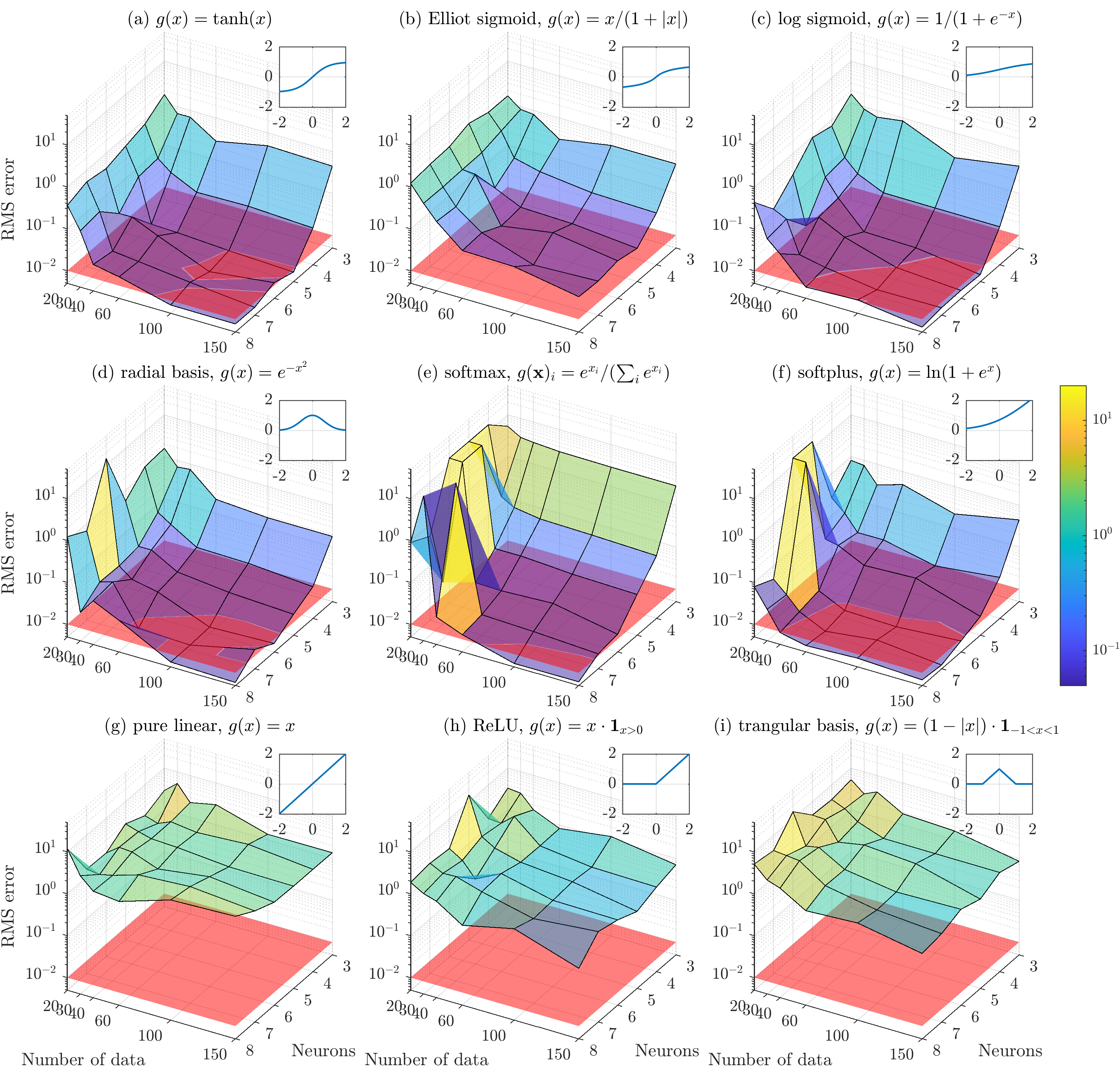}
\caption{\onehalfspacing RMS prediction error on testing data for different activation functions. The function forms are shown in the inset of each panel. The red transparent surfaces in each panel show a reference error level of $10^{-2}$. }
\label{fig:errors}
\end{figure}

\begin{figure}[ht!]
\centering
\includegraphics[width=0.6\textwidth]{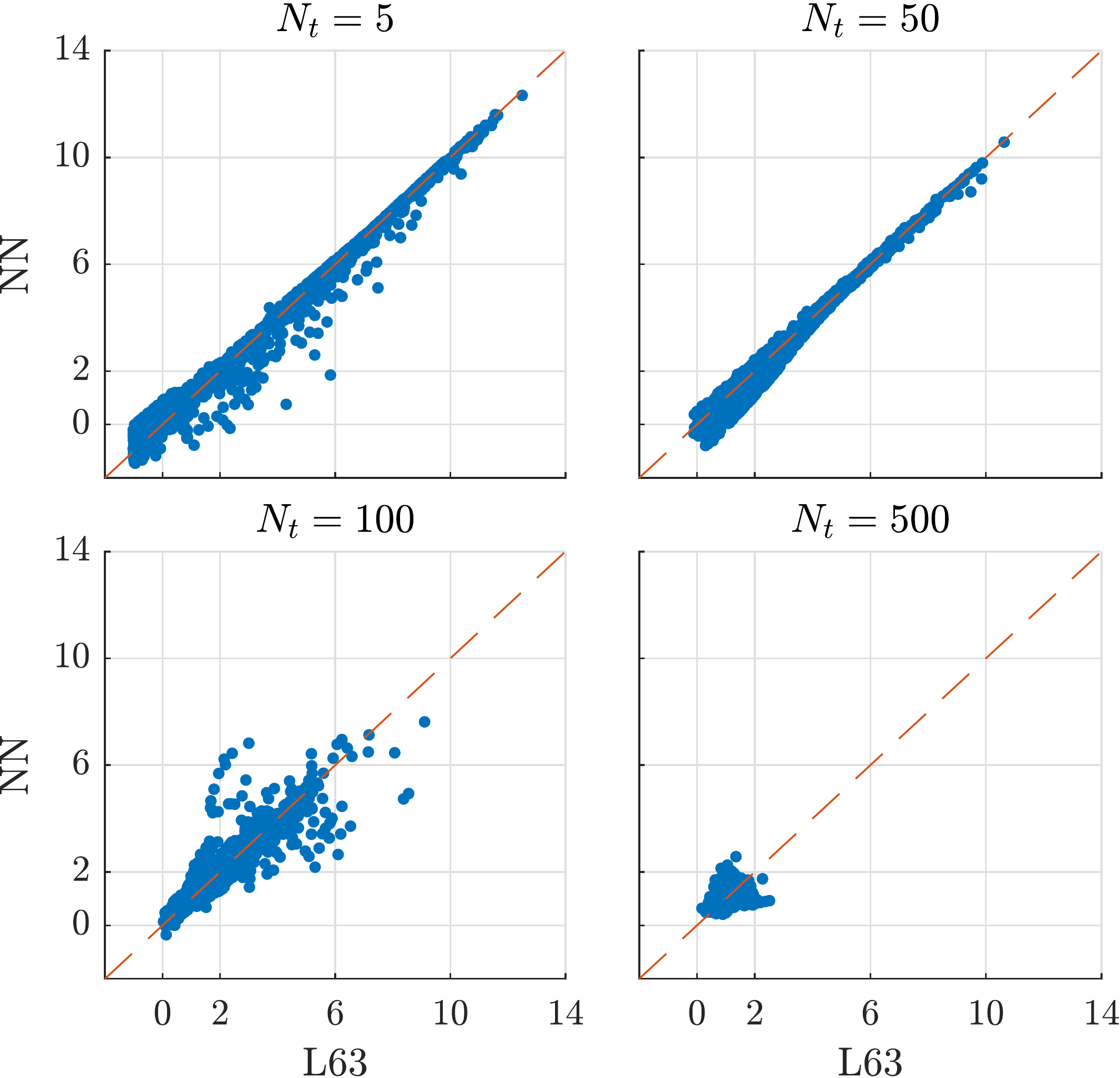}
\caption{\onehalfspacing Similar to 
%Fig.\;\ref{fig:FTLE}, 
Fig.\;2, 
but for the 5-neuron NN trained on 100 data points sampled from the $X>-5$ part of the L63 attractor.}
\label{fig:FTLE_half_domain}
\end{figure}

\begin{figure}[ht!]
\centering
\includegraphics[width=0.7\textwidth]{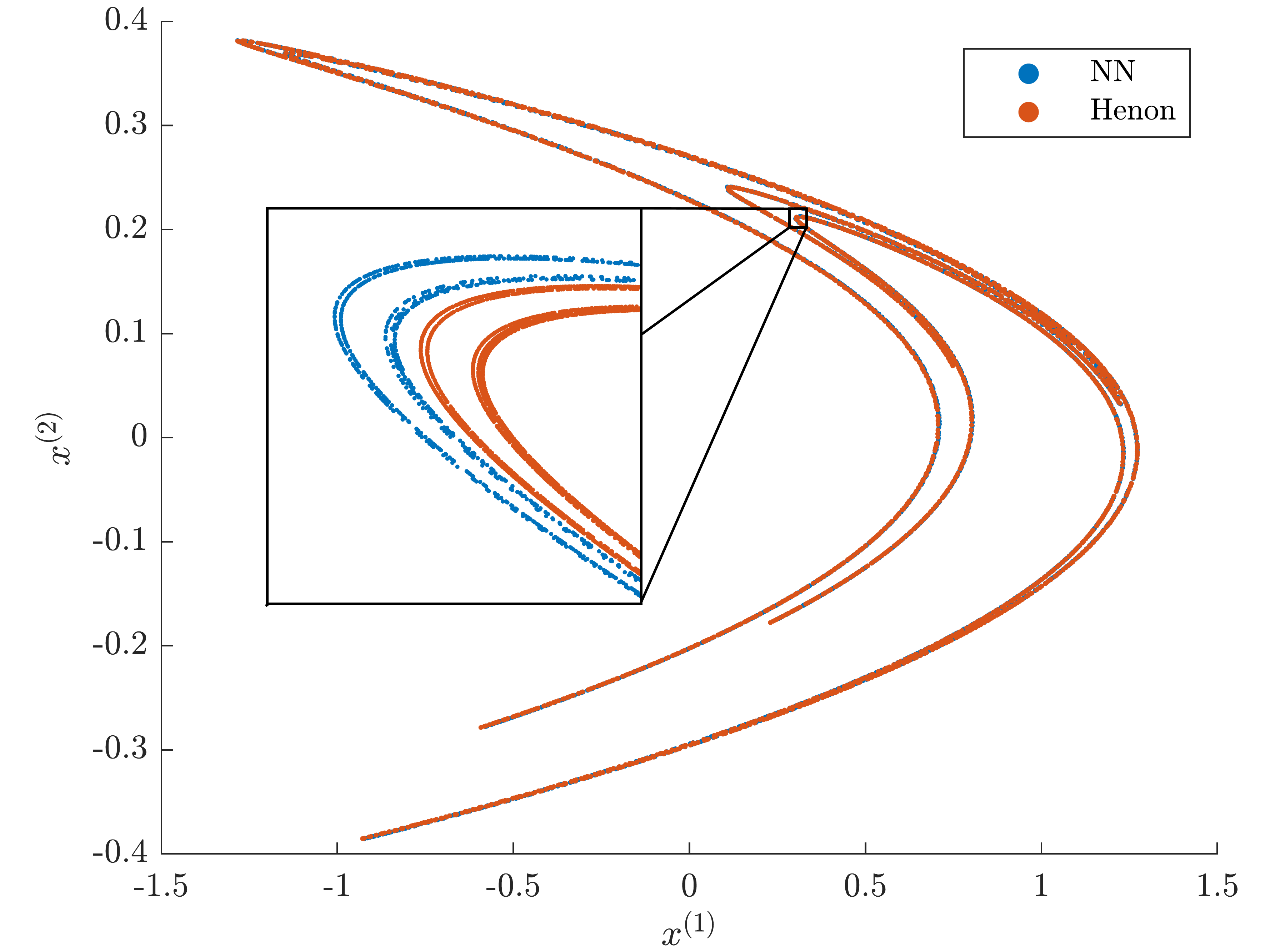}
\caption{\onehalfspacing Reconstructed attractor of the 2-neuron neural network trained on the H\'enon map. The inset shows a magnified region of $[0.295, 0.313]\times[0.206, 0.214]$, which shows the difference between the NN attractor and the H\'enon attractor. }
\label{fig:henon}
\end{figure}

%\begin{table}[!ht]
%\centering
%\caption{Exemplo de tabela de 3 colunas e 2 linhas}
%\label{tab:exTable1}
%\smallskip
%\begin{tabular}{l c c}
%\hline
%& Value 1 & Value 2\\[0.5ex]
%\hline
%&&\\[-2ex]
%Case 1 & 1.0 $\pm$ 0.1 & 1.75$\times$10$^{-5}$ $\pm$ %5$\times$10$^{-7}$\\[0.5ex]
%\hline
%&&\\[-2ex]
%Case 2 & 0.003(1) & 100.0\\[0.5ex]
%\hline
%\end{tabular}
%\end{table}

%\begin{lstlisting}

%int main(){
%  int a,b,c;
%  float x;
%  printf("informe o tamanho do lado do quadrado");
%  scanf("%d", &a);
%  printf("A area do quadrado %d", b=area(a));
%  printf("Duas vezes o valor do lado do quadrado %d", %c=aumenta(a));

%\end{lstlisting}